\documentclass[12pt]{article}
\usepackage{setspace}
\usepackage{amsmath}
\usepackage{fullpage}
\usepackage{arxiv}

\usepackage{mathpazo}
\usepackage{url}            
\usepackage{amsfonts}       
\usepackage{lipsum}
\usepackage{todonotes}
\usepackage[numbers]{natbib}

\def\keywordname{{\bfseries \emph Keywords}}%
\def\keywords#1{\par\addvspace\medskipamount{\rightskip=0pt plus1cm
\def\and{\ifhmode\unskip\nobreak\fi\ $\cdot$
}\noindent\keywordname\enspace\ignorespaces#1\par}}

\title{KnowBias: A Novel AI Method to Detect Polarity in Online Content}

\author{
  Aditya Saligrama \\
  MIT PRIMES/Weston High School \\
  Cambridge, MA 02139 \\
  \texttt{saligrama@csail.mit.edu} \\
}
\usepackage{setspace}
\date{}
\begin{document}
\maketitle

\begin{abstract}
We propose a novel training and inference method for detecting political bias in long text content such as newspaper opinion articles. Obtaining long text data and annotations at sufficient scale for training is difficult, but it is relatively easy to extract political polarity from tweets through their authorship; as such, we train on tweets and perform inference on articles. Universal sentence encoders and other existing methods that aim to address this domain-adaptation scenario deliver inaccurate and inconsistent predictions on articles, which we show is due to a difference in opinion concentration between tweets and articles. We propose a two-step classification scheme that utilizes a neutral detector trained on tweets to remove neutral sentences from articles in order to align opinion concentration and therefore improve accuracy on that domain. 
We evaluate our two-step approach using a variety of test suites, including a set of tweets and long-form articles where annotations were crowd-sourced to decrease label noise, measuring accuracy and Spearman-rho rank correlation. In practice, KnowBias achieves a high accuracy of 86\% ($\rho$ = 0.65) on these tweets and 75\% ($\rho$ = 0.69) on long-form articles. While we validate our method on political bias, our scheme is general and can be readily applied to other settings, where there exist such domain mismatches between source and target domains. Our implementation is available for public use at \texttt{https://knowbias.ml}.

\end{abstract}


\section{Introduction}
Rising bias in news media, along with the formation of filter bubbles on social media, where content with the same political slant is repeatedly shared, have contributed to severe partisanship in the American political environment in recent years \cite{renka_2010,kelly_francois_2018}. 

We aim to increase awareness of this heightened polarization by alerting users to the political bias in the content they consume. In this context we must impose the following constraints for any such system: 
\begin{enumerate}
    \item Our system should be \textit{generalizable} in that it can address bias in all types of text content without taking into account any additional features beyond the text itself, such as author or publisher. Consequently, we do not want to leverage user information such as browser history in order to address bias in the content they consume. 
    
    \item Additionally, our system should operate and provide information in \textit{real time}, requiring minimal bandwidth and data storage.
\end{enumerate}
These constraints necessitate a machine-learning approach in order to predict bias on new, changing data without using text metadata. We propose an NLP-based approach that predicts political bias on long text such as news articles independent of metadata such as content origin or authorship. Such supervised learning methods require a high volume of annotated training data. Annotating polarity on long documents at sufficient scale for training is infeasible since doing so requires that humans read each article and manually determine polarity. On the other hand, tweets can be easily gathered in high volume and can be annotated based on authorship.

We envision an approach where we transfer knowledge from tweets to long text at test time. While previous work has attempted to analyze tweets for political sentiment \cite{demszky2019analyzing}, there is no research on domain adaptation from short to long documents in this context. There has been research on filtering text for the purposes of deriving justifiable predictions \cite{lei2016rationalizing}, but not for domain adaptation for our target problem. Universal sentence encoders \cite{2018arXiv180311175C} provide good text representations regardless of target task and we would expect training a classifier on these to provide good performance on all text; however, this approach delivers inaccurate and inconsistent predictions. 

\begin{figure}[h]
    \centering
    \includegraphics[width=0.8\textwidth]{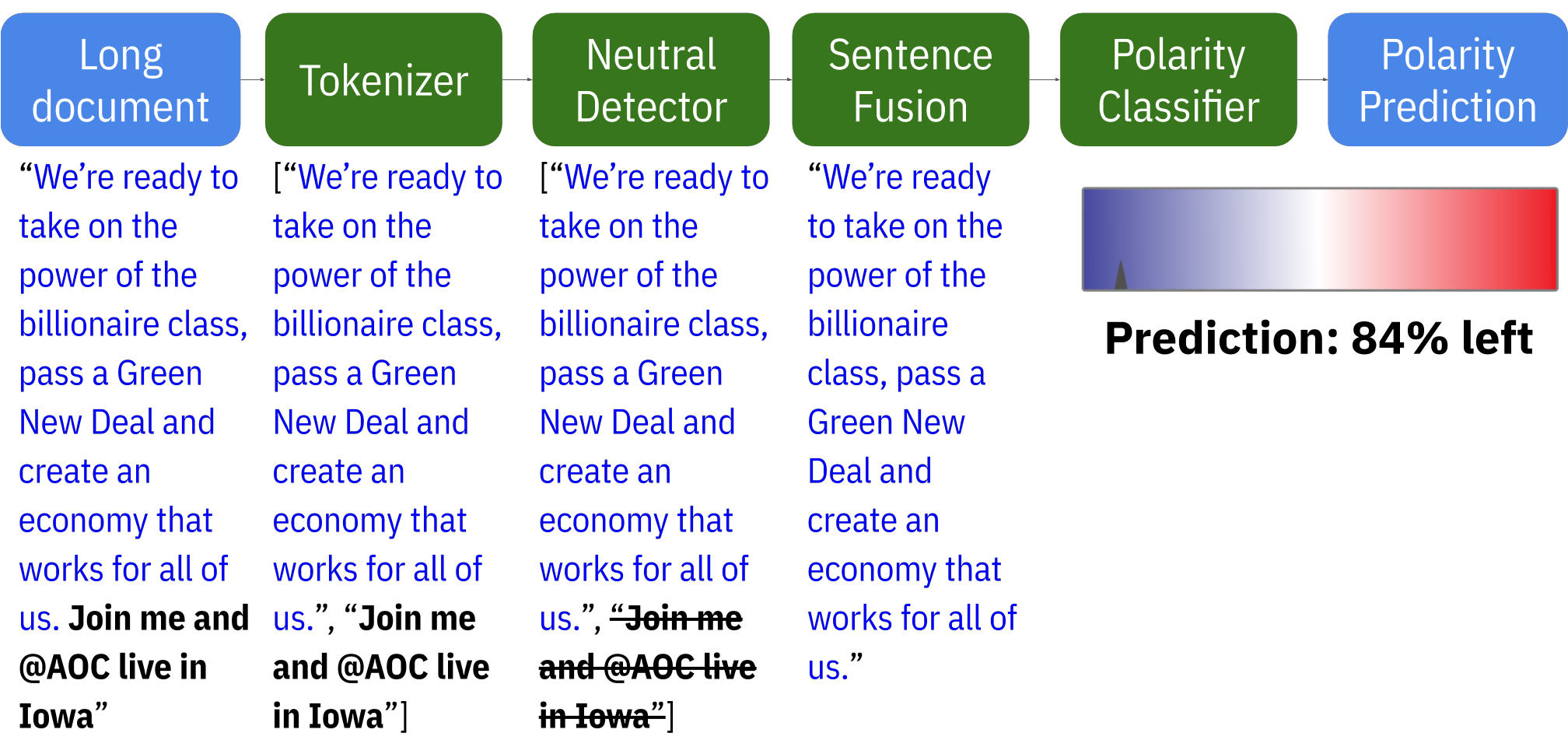}
    \caption{\small Proposed two-step classification scheme that tokenizes sentences in long documents and uses a neutral detector to filter out neutral sentences. Subsequently, it fuses remaining sentences to make a final prediction via a polarity classifier. Red sentences are polarized; black bold sentences are removed by the neutral detector.}
    \label{pipeline}
\end{figure}

We show that this poor performance is due to the existence of neutral, apolitical sentences in articles that dilute opinion concentration compared to tweets. Our proposed method alleviates this issue by using a neutral detector trained on tweets to remove neutral sentences before predicting bias, improving prediction accuracy and consistency. Our work extends \citeauthor{saligrama2019bknowbias} (\citeyear{saligrama2019bknowbias}).

\noindent \textbf{Other Applications.} Our work primarily focuses on the problem of detecting political bias. However, our method can be utilized for other tasks, such as analyzing rhetoric surrounding gun rights \cite{demszky2019analyzing} and social security \cite{dos2019classificaccao}, where research has focused on sentiment analysis on tweets. However, in these critical political issues, as well as in non-political realms such as sports \cite{sportsbias} or medicine,  long-form articles or publications can drive policy decisions. By training on short content such as tweets or abstracts and then transferring knowledge using our method, we can assess bias in such critical long-form content.

\section{Background}
We provide a brief summary of supervised learning, which consists of the following steps. 
\begin{itemize}
    \item Data Collection. Dataset in the form of annotated data ${\cal D}=\{(x_i,y_i),i=1,2,\ldots,N\}$ are collected. Here $x_i$ are the examples such as tweets, or text documents and $y_i$ are the labels such as polarity of the tweet or text document.
    \item Preprocessing. Examples are first converted into a feature representation for the purpose of convenient manipulation by a computer.
    \item Training. A classifier is trained to learn the relationship between inputs (text) and the desired output (polarity). 
    \item Inference. The classifier is then test set to predict outputs for heretofore unseen examples.
\end{itemize}
From a mathematical perspective, machine learning models can be expressed as a function $f(x_i) = p_i$ where $x_i$ is an input example, $p_i$ is a class prediction for that example, and $f(\cdot)$ is a classifier. In our specific case, we use a form of machine learning called representation learning \cite{2012arXiv1206.5538B}, where the model can learn feature representations from training data, which are necessary to discriminate raw data into two classifications.

We train this type of model, such as a neural network, on a set of training data in a supervised learning scheme, which is a subset of representation learning. A neural network consists of many layers. Each layer, $t$, is parametrized by a matrix of weights $w(t)$ and biases $b$, and the output of the subsequent layer is obtained by a recursion: $x(t+1)=\phi(w(t)x(t)+b)$, with the first layer $x(0)$ being the input $x$. Here $\phi(\cdot)$ is an activation function. Different activation functions have been used in practice. 

The final layer, $T$ is composed of a vector of weights $w$ and a vector of biases $b$, with length being the number of features in the input data. A prediction probability is determined using a sigmoid function
\begin{align}\label{sigmoid}
    p(x(T), w(T), b) = \sigma(w(T) x(T) + b) = \frac{1}{1 + e^{-(w(T) x(T) + b)}}.    
\end{align}
Note that $x(T)$ is a function of $x(T-1)$ and which is in turn a function of $x(T-2)$ and so on. The overall model $f(\cdot)$ is equivalent to $p(x(T), w(T), b)$, from which binary predictions are obtained, and we output a label that corresponds to the class with higher probability.

The set of all parameters $w, b$ is denoted by the parameter vector $\theta=(w_1,w_2,\ldots, w, b)$. 
During training, mispredictions on the training set are penalized using a log-loss function
\begin{align}\label{logloss}
    \ell(p_i, y_i) = -(y_i\log{p_i} + (1-y_i)\log{(1-p_i)}).
\end{align}
Here, $y_i$ is the ground-truth classification label for example $i$ and $p_i$ is the predicted probability that example $i$ is classified as $y_i$. This loss function penalizes all errors, but most severely those with a high probability corresponding to the wrong classification. As the probability of misprediction for example $i$ rises, $\ell(p_i, y_i)$ goes to infinity, while as the probability of a correct prediction rises, $\ell(p_i, y_i)$ goes to zero.

The overall loss function for a model $f_{\theta}$ on a dataset ${\cal D}$ is:
\begin{align}\label{overall_logloss}
    L(f, {\cal D}) = \sum_{i=0}^{N} \ell(f(x_i), y_i),
\end{align}
\noindent where $N$ is the number of examples in the training set. Training attempts to minimize $L(f, {\cal D})$ using stochastic gradient descent \cite{ng_ngiam_foo_mai_suen_coates_maas_hannun_huval_wang_et_al}, in which a derivative of the loss function is taken and model parameters of $\theta$ are adjusted in the opposite direction of that derivative. This can be visualized as taking a step down the gradient towards a minimal loss. The size of this step, $\eta$, is a predetermined parameter called the learning rate. Concretely, suppose the dataset is given by ${\cal D}=\{(x_i,y_i), i=1,\,2,\,\ldots,N\}$. The stochastic gradient algorithm takes a random sample from the dataset ${\cal D}$ and iterates as follows:
$$
\theta^t \leftarrow \theta^{t-1} - \eta \nabla \ell(f_{\theta}(x_i),y_i)
$$

\noindent{\bf Preprocessing for Natural Language Processing (NLP).} In NLP, directly using phrases as the training data does not work very well as sentences are not machine-interpretable. A number of methods have been researched that convert natural language into a format that can be understood by computers. Word embeddings \cite{2013arXiv1310.4546M} are currently the most popular such representation, because they can capture semantic meaning in words.

\noindent{\bf Word Embeddings.} These embedding suites generally act as functions $g(x) = \mathbf{v} \in \mathbb{R}^D$ that convert words $x$ into high-dimensional vectors $\mathbf{v}$ in a D-dimensional real space. Recent research has yielded many different such embedding suites. Some of these models are bag-of-words models in that they do not consider the order of words in a sentence, whereas more recent developments in sentence encoders can utilize word order to produce more accurate embeddings. 

A key reason NLP researchers are excited about word embeddings is that 
\begin{enumerate}
    \item A vector representation that accounts for context of a word in a document, semantic and syntactic similarity, and relation with other words. 
    \item The feature representation can be used as an input for training on new tasks, which we do not need to know beforehand. Namely, it allows for transferring knowledge to new tasks without requiring learning new feature representations. 
\end{enumerate}
Such vector representations have had much success in many NLP applications. A common illustration of the utility of word embeddings is to solve analogy tasks based on mapping the analogy task into a linear space and using linear algebra to determine the solution to the analogy. For instance, the analogy \texttt{king:queen::man:?} can be solve by first mapping words into a high-dimensional vector space and obtaining vectors $v_{king}, v_{queen}, v_{man}$ and looking for a word $w$ such that the angle between the vectors $v_{king}-v_{queen}$ and $v_{man}-v_{w}$ is the smallest. This is known as the cosine similarity. 

\section{Related Work} \label{sec:related_work}
The issue of analyzing text for bias of all types has been researched in social science fields as well as in the broader machine learning (ML) and natural language processing (NLP) spaces. There are a number of methods that detect polarization in text and media \cite{gentzkow_shapiro_2006, gentzkow_shapiro_taddy_2016}. These works propose generative models to elucidate properties for specific datasets such as congressional records and describe methods for prediction within the same context (i.e., training and testing on congressional records). However, congressional records are short and not representative of long-form articles, encountering a similar domain-adaptation problem.

Recommendation systems \cite{diresta_2018} provide readers with curated content. However, these engines utilize  reader profile data and browsing history, along with rule-based tables for identifying left/right spectrum of online content to tailor their recommendations. As such, a recommendation-engine solution is not generalizable and fails to meet our criteria for a solution.

There is extensive literature \cite{grimmer_stewart_2013} in the social science literature on polarization that deals with mathematical models and supervised learning techniques to accurately identify polarity in written content. While the survey recognizes the need for scalability, the authors are pessimistic that automatic content analysis can fully replace humans. Moreover, the methods described in the survey hinge on the availability of reliable and clean annotated data, which we do not have access to in our problem. 

A number of works in ML and NLP literature propose methods for sentiment analysis. In particular, domain adaptation is a recurring theme \cite{Glorot:2011:DAL:3104482.3104547, Socher:2011:SRA:2145432.2145450} and this aspect is indeed closely related to our problem. In this context, several learning methods leveraging universal word-encoders networks, which are increasingly gaining importance, have been proposed. Universal encoders have been shown to be effective in mapping words (or phrases and sentences) into high dimensional Euclidean vectors while preserving their semantic and contextual meaning, while being agnostic to target task \cite{2018arXiv180311175C, 2013arXiv1310.4546M, Pennington14glove:global}. These vectors serve as an input for supervised classification for any downstream task. 

Nevertheless, the emphasis in existing works is on short texts, and in particular, domain adaptation \cite{Glorot:2011:DAL:3104482.3104547, Socher:2011:SRA:2145432.2145450} scenarios described deal with similar tasks and contexts during training and testing. For instance, experiments described in \cite{Socher:2011:SRA:2145432.2145450} predict the sentiment of kitchen reviews by transferring knowledge from furniture reviews. No prior work in the domain-adaptation space exists for training on short-form content such as tweets and testing on long-form article data.

Recently, several works have highlighted the importance of using a neutral class as a third class in sentiment analysis \cite{vryniotis_2013, Schler05theimportance}, which could potentially be useful for our problem as well. However, these methods again require well-annotated training data and assume similarity between training and target content. 

In summary, while existing concepts and tools from ML and NLP can be useful, there is no existing work that directly deals with our proposed problem.
\section{Predicting Polarity in Text Content}

\begin{figure}[t!]
    \centering
    \includegraphics[scale=0.47]{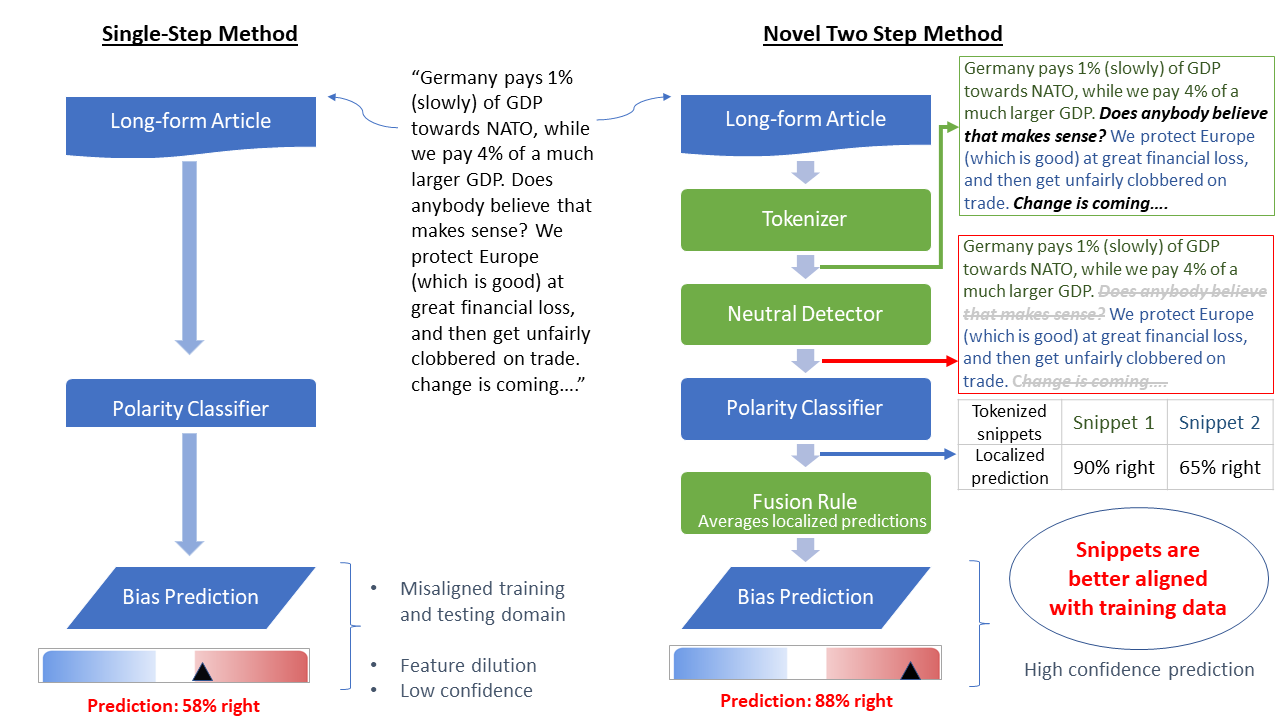}
    \caption{Single vs. Two-Step Classifier Scheme for Polarity Classification. In the single-step scheme (TEPC), we simply use an encoder suite (WWE or LUSE) to convert articles into vectors and then use a classifier trained on tweet vectors to detect polarity. Due to stylistic differences between tweets and long-form articles, TEPC predictions on long-form articles were inaccurate and unconfident. However, with the two-step scheme, we first tokenize articles into sentences, and use a neutral detector to filter out neutral sentences. The remaining sentences, which have a similar opinion concentration to tweets, are fused together and a TEPC is run to obtain a final prediction.}
    \label{pipeline}
\end{figure}

We proceed with a supervised learning approach in order to detect polarity. As discussed in Section~\ref{sec:related_work}, a supervised learning problem is composed of: 
\begin{itemize}
    \item[a.] Data collection: We collect examples and corresponding ground-truth annotations denoting the polarity/bias of the example. In our work, ground-truth annotations are generally subjective, since polarity rating varies across different people. 
    \item[b.] Preprocessing. As is standard in natural language processing works we experimented with different preprocessing procedures including tokenization, substitution, normalization, lemmatization etc.
    \item[c.] Universal Feature Representation. We map pre-processed textual data after tokenization into vector spaces that preserve semantic and contextual information. In this context we leverage recent works that show that universal encoders are effective in mapping words (or phrases and sentences) into high dimensional Euclidean vectors while preserving their semantic and contextual meaning, while being agnostic to target task \cite{2018arXiv180311175C, 2013arXiv1310.4546M, Pennington14glove:global}. We experiment with different encoders. 
    \item[d.] Training. We then train a Neural Network for predicting polarity or bias on training data. We experiment with different architectures. 
\end{itemize}
We will now discuss these main steps in the following subsections.


\subsection{Data Collection}

We train on political tweets due to the aforementioned ease in collecting and annotating them at scale and aim to transfer this knowledge to longer articles. Our polarity data consisted of roughly 150,000 tweets from 28 Twitter verified politicians or media personalities across the political spectrum. 80\% of these samples were used for training and 20\% were used as testing. We also sampled a set of roughly 80,000 neutral tweets from the Twitter general stream in order to train the neutral detector. Figure \ref{example_tweets} illustrates example tweets used for training.

\begin{figure}[t!]
\small
\textbf{Example Highly Left-Biased Tweet}: \texttt{I hope that we can move swiftly to conclude this discussion about party positions, so that we can spend more time discussing party priorities: voting rights, healthcare, wages, climate change, housing, cannabis legalization, good jobs, etc.}
\newline
\textbf{Example Highly Right-Biased Tweet}: \texttt{Sure, let this migrant caravan come on in. From then on every registered Democrat should be responsible for paying for every single expense they cost the rest of the American taxpayers. You want em? You pay for them. \#MigrantCaravan}
\newline
\textbf{Example Neutral (Noisy) Tweet}: \texttt{Let's not ignore the fact that how beautifully Hobi sang this part of Spring Day!! His voice is so soothing!!}

\caption{Example Twitter phrases acquired in late 2018 for training and testing. Neutral tweets were used to train the neutral detector that can discriminate and filter out apolitical phrases.}
\label{example_tweets}
\end{figure}
\subsection{Baseline method}
Our baseline method features a simple deep-neural-network (DNN) classifier \cite{scikit-learn} which takes word embeddings as input and outputs political polarity. We call this model the Text Embedding and Polarity Classifier (TEPC). The fact that universal encoders can extract semantic meaning agnostic to target task makes this approach feasible.

The DNN classifier is trained on fully annotated Twitter data. We use one of the encoders listed on Google TensorHub \cite{ww_500, guse_2, guse_large_3} to output these word embeddings, which are usually vectors between 100 and 512 dimensions. We tested modifications such as max-pooling of word or sentence features but eventually settled on the Google Universal Sentence Encoder Large \cite{guse_large_3} (LUSE) as the empirically optimal encoder. We also performed 5-fold cross validation with an 80-20 train-test split to grid search and choose the best parameters for the learning rate, regularizers, number of layers, and hidden nodes for the DNN classifier.

We found that while our baseline method performed well on the same domain, namely, trained on Twitter and tested on Twitter (see Table~\ref{tepc_results}), it performed poorly on cross-domain tests (trained on tweets and tested on Reddit data as well as long form articles). This was not entirely surprising. We did not anticipate that this method will perform well in domain adaptation-reliant test cases (i.e., long-form articles) as it relies on the assumption that test data is similar to the Twitter training data. Instead, this method serves only as a baseline to which we can compare an improved approach for domain adaptation scenarios. Our goal is to allow for polarity prediction on long-form articles.
\subsection{Opinion concentration and neutral sentences} 
Long-form articles have a mix of neutral and political sentences, while tweets are concentrated with opinion. A long-form article typically contains a large number of sentences that are somewhat neutral. These sentences are usually long, descriptive, and do not contain much political languages. We observe that performance degrades on longer paragraphs with larger number of neutral sentences.

If our insight holds, then one way to improve accuracy is to take a long-form article, tokenize it, then filter-out seemingly neutral sentences, and then predict bias sentence-by-sentence. Finally, we use a fusion rule to combine sentence-level predictions to output a final terminal prediction.

As a consequence, we proceed as follows:
\begin{itemize}
    \item Test whether neutral sentences significantly dilute content and degrade prediction.
    \item Determine whether detecting a neutral sentence is relatively easy.
\end{itemize}

\subsection{Embedding Dilution on Long-Form Articles}
We test the hypothesis that longer texts degrade performance regardless of encoding method or classifiers. We create the synthetic Twitter Concatenated dataset that augments each test sentence in the Twitter Political dataset with a random neutral sentence. We study the effect of dilution by increasing the number of augmented sentences from 1 to 5 for each tweet.

\begin{figure}[t!]
    \centering
    \includegraphics[scale=0.4]{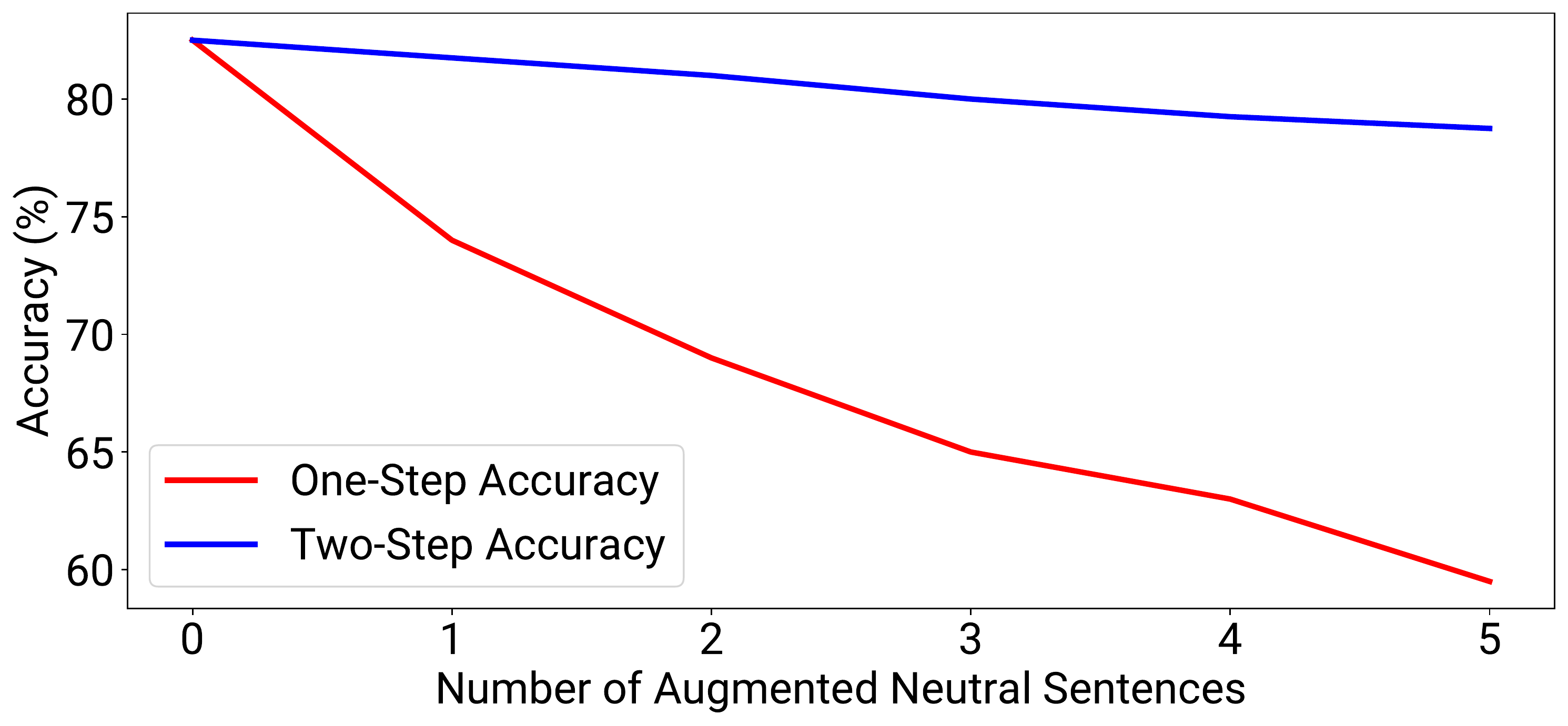}
    \caption{Degradation of accuracy after neutral sentence augmentation with TEPC vs. Two-Step classification approaches. The Two-Step method degrades gracefully relative to the TEPC method as a result of removal of augmented sentences by the neutral detector. Although LUSE embeddings were used for this experiment, a similar effect can be observed when using WWE embeddings.}
    \label{degradation}
\end{figure}

We then test prediction degradation with trained TEPCs tabulated in Table \ref{tepc_results}. Note the significant degradation in TEPC accuracy in Figure \ref{degradation}.

\subsection{High Accuracy Neutrality Detection}

Given that degradation occurs when neutral sentences are augmented, it is clear that filtering out these neutral sentences can improve accuracy. This strategy can only be effective if we can detect neutral sentences easily. Surprisingly, testing on the Twitter Concatenated dataset (see Table \ref{tepc_results}) shows that this is indeed the case, where nearly 95\% accuracy using any encoder suite.

To build intuition, we examine explained variance ratios (EVR) of principal-component analysis (PCA) \cite{Abdi:2010:PCA:3160436.3160440} for 500 randomly sampled differences of left and right LUSE vectors on the Twitter Political dataset. We compare these EVRs of neutral and biased vectors from the Twitter Concatenated dataset. Results for EVRs show that most of the energy for neutral-bias vectors is concentrated in one component while the energy for left-right vectors is spread out among many components. From this we infer that the neutral-bias data can be clustered into a neutral cluster and bias cluster separated by a hyperplane. We therefore train a DNN on LUSE vectors on the Twitter Concatenated data and realize an accuracy of over 95\%.

\begin{figure}[t!]
    \centering
    \includegraphics[]{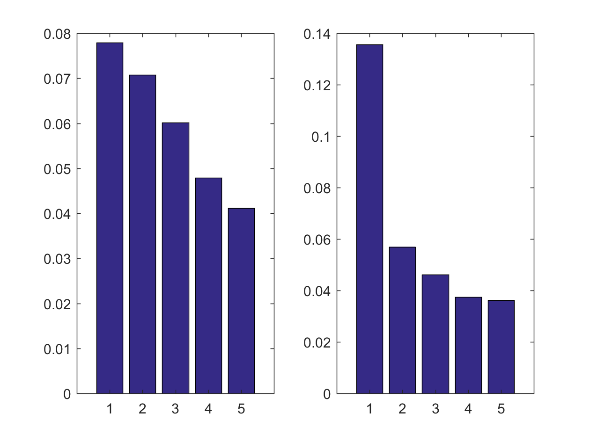}
    \caption{Explained Variance Ratios for Left-Right encoded vectors (left) and Bias-Neutral vectors (left) output by principal component analysis (PCA). While the differences between left and right vectors are spread out over many components and therefore more difficult to separate using a classifier, the differences between bias and neutral vectors are largely concentrated in one component and therefore these vectors are easy to separate. This suggests that effective neutral detection is possible.}
    \label{pca-evr}
\end{figure}

\subsection{Redesigned Approach for Domain Adaptation}

Figure \ref{pipeline} outlines our method, comparing the baseline single-step method versus the novel two-step method that better overcomes the challenge of domain mismatch.

To overcome the issue of domain mismatch, we redesign our method to better align training data with target domains such as long-form articles. As shown in Figure~\ref{degradation}, accuracy is significantly hampered by the presence of neutral sentences, so we use our trained neutral sentence detector to distinguish and remove these sentences as follows.


%

We propose a two-step cascade classifier scheme where we use two different DNNs to overcome domain mismatch. We first split test data into sentences and use the neutral detector to filter out neutral sentences. Subsequently, we merge the remaining polar sentences and predict a final polarity score on this corpus.

In the following subsections, we detail each of the components of our cascade scheme.

\subsubsection{Neutral Detector}

After identifying the dilution of opinion concentration as responsible for accuracy degradation on long-form articles, we propose the addition of a classifier to detect and remove neutral sentences. We train a second deep neural network on the sentence embeddings of 80,000 tweets sampled from the general Twitter stream as well as the political samples, obtaining a high 95.63\% accuracy. 

\begin{figure}[h]
    \centering
    \includegraphics[scale=0.3]{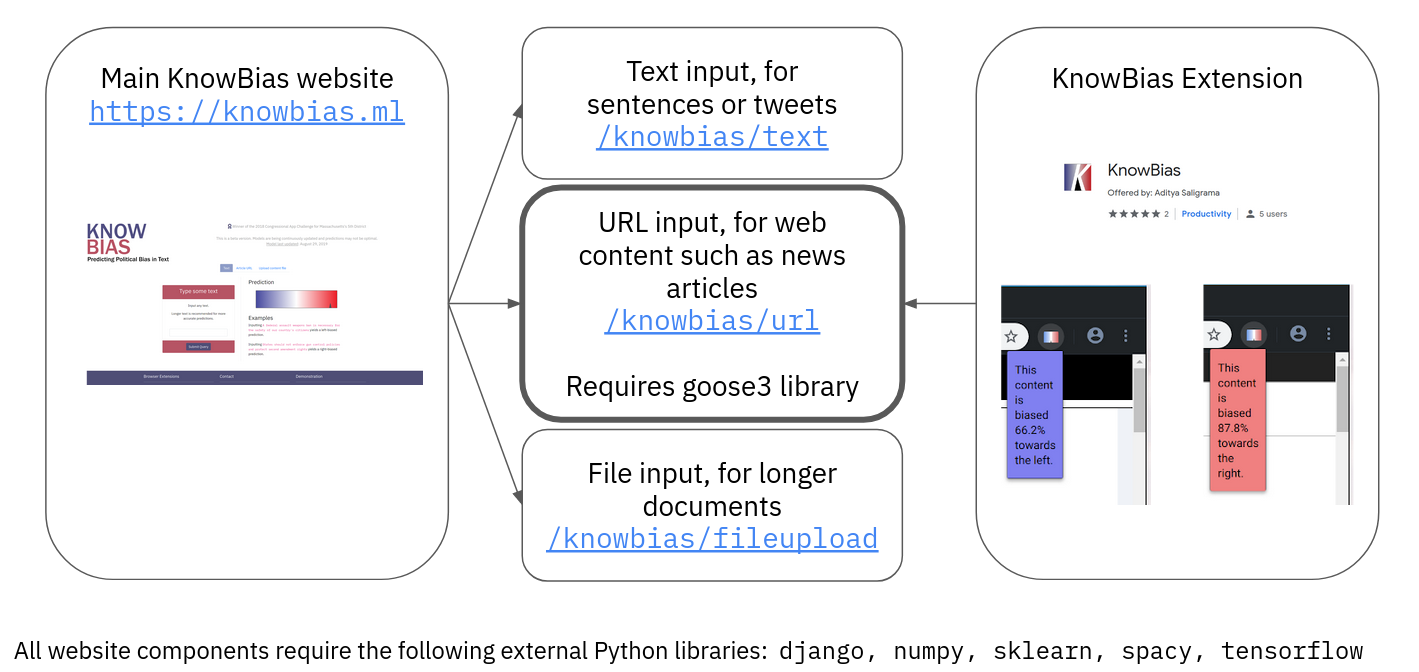}
    \caption{The browser extension is written in JavaScript using WebExtensions technology, which allows identical code to be used to create Firefox and Chrome extensions. When the status of the active browser tab is changed (i.e., the user navigates to a new site in that tab or a different tab is clicked on), an event fires. When that event is detected, the extension retrieves the active tab's URL and sends a GET request to the URL section of the website, which returns a prediction. This prediction is kept in the extension's local storage and read by the display section of the extension when the active tab change event is fired, displaying the new prediction. The browser extension involves minimal data overhead for users and maintains user anonymity.}
    \label{interface}
\end{figure}

\subsubsection{Polarity Prediction with Two-Step Cascade}

We propose a two-step classification scheme in order to improve prediction quality on long-form articles as demonstrated in Figure \ref{pipeline}. On any data passed to the system for inference, we first tokenize it into individual sentences. On each of these sentences, we use the neutral detector to mark and remove all neutral sentences. We then fuse the remaining sentences back together, aligning opinion concentration to that of tweets, and then use the main baseline classifier to predict polarity.

\subsection{Frontend Interface}
We additionally provide an efficient front end with a light footprint and low latency that is compatible with any browser. 
The website is available for public use at \texttt{https://knowbias.ml}, where users can enter example text, upload a plaintext content file, or submit an article URL for polarity evaluation. For example, entering the phrase \texttt{Americans' Second Amendment rights must be protected} yields a center-right biased prediction. Links to browser extensions for all Chrome-based and Firefox-based browsers are provided from the website. Because only the content URL is communicated from the website and extensions to the computation server, the data overhead for the user is minimal, as demonstrated in Figure \ref{interface}. When asked to provide usage feedback on the front end, users did not observe unusual delays in different test-case scenarios, such as on mobile and desktop platforms.

\section{Experiments}

We describe the different test suites that we evaluated KnowBias on, testing both our baseline TEPC and the novel two-step cascade approach. Experiments were run using Google Colaboratory due to the generous computing resources provided by the platform.

\subsection{Polarity Detection on Twitter Test Data}

Table \ref{tepc_results} tabulates the results from testing different encoders for the baseline TEPC. Overall, we found that using LUSE \cite{guse_large_3} coupled to a DNN with 2 hidden layers realized the best performance  on all datasets, achieving 82\% on the Twitter Political dataset. However, wiki-word embeddings (WWE) are significantly faster in terms of runtime and memory use, since they are word-based and ignore punctuation, while being comparable in performance to other encoder suites.

\begin{table}[t!]\label{tepc_results}
\centering
\fontsize{10}{11}\selectfont
\begin{tabular}{|l|l|l|l|l|l|}
\hline
                            & \shortstack{Wiki \\ Words \cite{ww_500}} & \shortstack{Universal\\ Sentence \\ Encoder \\ Small \cite{guse_2}} & LUSE \cite{guse_large_3}  & \shortstack{Max-Pool \cite{2018arXiv180509843S} \\ Wiki Words} & \shortstack{Norm1-128 \\ \cite{norm1_128}} \\ \hline
Twitter Political           & 77.89      & 77.50      & 82.27 & 74.50               & 76.00     \\ \hline
Twitter Concatenated        & 94.20      & 93.00      & 95.63 & 94.50               & 95.11     \\ \hline
Reddit Political            & 72.30      & 73.10      & 75.50 & 72.43               & 70.00     \\ \hline
Cross-domain Twitter-Reddit & 60.01      & 60.12      & 62.02 & 62.32               & 62.01     \\ \hline
\end{tabular}
\newline
\caption{TEPC test results across various encoder suites for four datasets obtained by 80-20 train-test split. Twitter Political is the 20\% split from the TEPC training data collected up to November 2018. Twitter Concatenated is the Twitter Political set plus tweets taken from the general stream collected up to November 2018 (i.e., the 20\% test split from the neutral detector's training data), where political tweets are labeled as ''biased'' and noise tweets are labeled as ''neutral.'' Reddit Political is data from four highly polarized Reddit subreddits collected up to May 2018. Cross-domain Twitter-Reddit data is a combination of the Twitter Political and Reddit Political data. A DNN classifier sourced from Scikit-Learn \cite{scikit-learn} was used in all cases.}
\end{table}

\subsection{Two-Step Cascaded Classifier}

We conducted several experiments to validate accuracy of the two-step cascaded classifier. First, we considered the synthetic dilution experiment and applied a TEPC on text samples filtered by the neutral detector. Figure \ref{degradation} shows that there is negligible degradation in accuracy with augmented neutral sentences for the proposed approach, in contrast to applying a TEPC on the original text.

\subsection{Crowdsourced Annotation Study}

\begin{figure}[t!]
    \centering
    \includegraphics[scale=0.4]{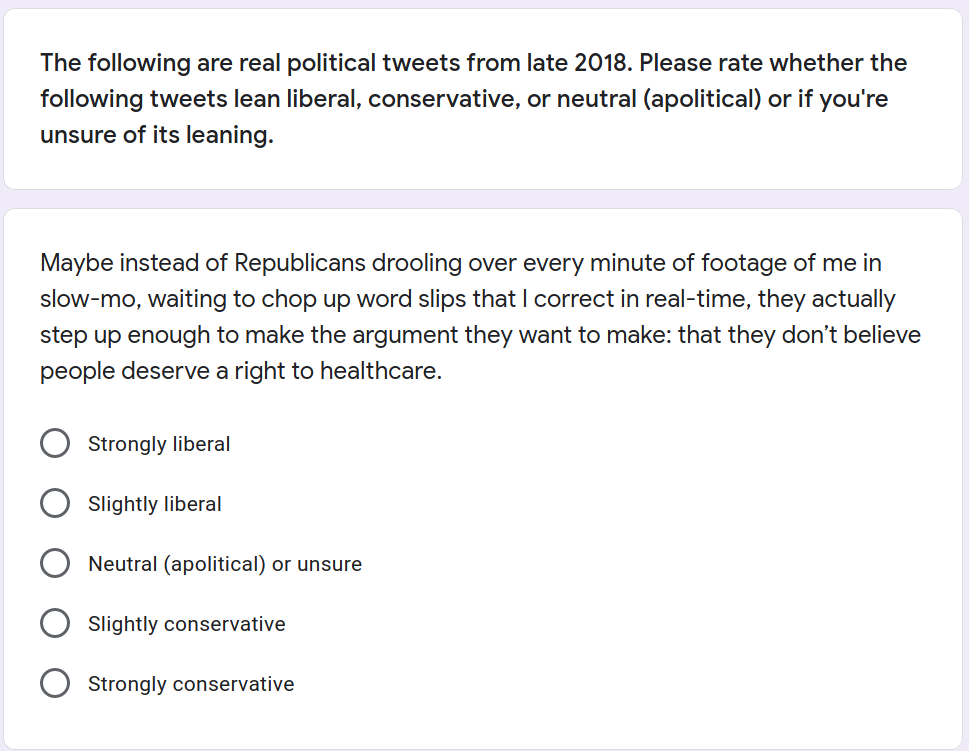}
    \caption{Example annotation survey sent to users of several Discord groups and the author's social media followers. Five of these surveys were created and a website randomized which survey a user would answer. In total, we collected 79 responses across a variety of user political affiliations.}
    \label{survey}
\end{figure}
To validate prediction accuracy, a smaller study was run, using politically knowledgeable users to crowdsource annotations for a set of tweets. The author surveyed several Discord groups and his social media followers, creating 5 Google Forms each with 10 tweets \cite{survey}, and recorded annotation responses. Respondents were asked to classify as ''strongly conservative,'' ''slightly conservative,'' ''neutral (centrist) or apolitical,'' ''slightly liberal,'' or ''strongly liberal'' as well as to state their personal political affiliation on the same spectrum. 79 users responded to the surveys, with a nearly balanced set among different political affiliations. Accuracy on these tweets with crowdsourced annotations was 86\%, which was higher than accuracy on the Twitter Political dataset (Table \ref{tepc_results}). We did not expect our accuracy to substantially change on the Twitter data between the TEPC and two-step methods, and indeed this is the case with no improvement in accuracy.

In parallel, we distributed 24 articles from various left and right-leaning news outlets. Five workers among the author's social media followers responded and their opinions were collected. We obtained an accuracy of 66\% with the single-step system vs. 75\% with the two-step cascade classifier.

Furthermore, to check the consistency of our polarity ranking, we also computed Spearman-rho rank correlation \cite{mcdonald_2015} between system predictions and crowd opinions. This statistical test evaluates how well system annotations correlate to human annotations with respect to the relative percentage of polarity. Specifically, in our context, it is useful because it measures the consistency of machine predictions with that of the humans while being agnostic to exact prediction values of our AI system. The Spearman-rho correlation is computed based on correlating the ranking provided by humans for each article against that predicted by our AI system, namely, 
\begin{align*}
    \rho = 1 - \frac{6 \sum_{i=0}^N (r_{h_i} - r_{m_i})^2 }{N^3 - N}
\end{align*}
where the $r_{h_i}$ are human ranks and $r_{m_i}$ are machine ranks, and $N$ is the amount of test data. Values close to -1 indicate a highly negative correlation, while values close to 1 indicate a highly positive correlation. Results tabulated in Table \ref{spearman} show that the proposed system for both long and short articles not only matches crowd opinions on average, but is also consistent in assigning the degree of polarity due to the relatively high Spearman rank correlation coefficient. The two-step system ($\rho = 0.69$) is far more consistent in assigning predictions with respect to crowdsourced annotations on articles than the baseline one-step method ($\rho=0.52$).

\begin{table}[t!]
\label{spearman}
\centering
\begin{tabular}{|l|l|l|l|l|l|}
\hline
Task                     & TEPC  & Two-Step \\ \hline
Twitter Political - Accuracy          & 82.27\% & 82.42\%    \\ \hline
Twitter Crowdsourced - Accuracy       & 86.00\%  & 86.00\%   \\ \hline
Twitter Crowdsourced - Spearman-{$\rho$}   & 0.65  & 0.65    \\ \hline
Articles Crowdsourced - Accuracy           & \textbf{66.67\%}  & \textbf{75.00\%}   \\ \hline
Articles Crowdsourced - Spearman-{$\rho$}       & \textbf{0.52}  & \textbf{0.69}    \\ \hline
\end{tabular}
\caption{\small Accuracy and Spearman-rho Rank Correlation \cite{mcdonald_2015} of TEPC and the two-step cascade classifier on tweets and long-form articles.}
\end{table}

\section{Conclusions and Future Work}
We introduced a real-time machine learning system to detect polarity in text content that does not rely on user data or content metadata to provide predictions. We demonstrated that it is possible to overcome the issue of domain mismatch between tweets and long-form articles by using a two-step approach to filter out neutral, descriptive sentences in the latter and better align the two domains by equalizing opinion concentration. 

Future work on KnowBias may involve exploring the problem of time shift, where positions on new issues are not accurately represented by predictions if the training data is too stale. This reinforces the need for continuous model updates. Additionally, the model could benefit from better data collection in terms of acquiring better neutral data, as we noticed that a few false negatives as well as several false positives appeared in our filtered list.

To conclude, the literature survey \cite{grimmer_stewart_2013} is skeptical of the ability of mathematical models to predict polarity in text. The meaning of words change over time and common literary devices such as sarcasm, irony and rhetoric are difficult for a model or machine to decipher. Nevertheless, this work shows that it is possible to build a relatively simple and efficient system whose predictions are consistent with human judgement as long as the system is periodically updated with more recent data. While there is no substitute to careful analysis and close reading that only a human can do, our proposal can serve as the first step of awareness for users.

\section{Acknowledgements}
I would like to acknowledge Prof. Kai-Wei Chang (UCLA) for feedback and comments on this paper, and for providing significant commentary on how the method described in this work could be utilized in other applications. Early into the project, he also helped me brainstorm different methods for data collection and architectures for sentiment analysis.

I would also like to thank Jon Gjengset and Prof. Frans Kaashoek (MIT), my 2018 MIT PRIMES mentors, for their guidance in defining a focus for KnowBias early on. In general, I have learnt and developed much intuition from my time in MIT PRIMES, including my experience working with adversarial machine learning at Madry Lab. 

I would like to thank my computer science teacher, Ms. Alison Langsdorf for listening to my ideas and suggesting me many possible next steps. She has always made time in her busy schedule to help me brainstorm ideas and to consistently encourage me to push forward.

My AP U.S. History teacher, Mr. Sean Smith, and my sophomore English teacher, Mr. Matthew Henry, have shared in my excitement as I built KnowBias. They have shared my website with their students, whom I have been able to collect feedback from to improve the tool.

Prof. Margo Seltzer (UBC), with whom I have worked on several other projects, has been a patient sounding board as I wrestled through data challenges and early frustrations. Working on the Certifiably Optimal Rule Lists interpretable machine learning project under her guidance inspired me to turn an eye towards other uses of machine learning for fairness, as demonstrated with KnowBias.

I also would like to thank Emma Hsiao for creating the KnowBias logo and helping with some of the graphics for the website, and Neil Malur, Anthony Cui, and Josh Yi for help with early teething issues with the code.

Since I was testing the accuracy of my model for which there is no ground truth, I crowdsourced data from several sources to serve as annotations for one of the experiments. I would like to thank my friends and users of several Discord groups I am involved in, and my followers on Facebook and Instagram, who responded to my annotation surveys instantly. In just a few days, I was able to get 79 responses on this survey.

Finally, I am very appreciative of and grateful to my family for their love, patience and support, without which this project would not have been possible.

\bibliographystyle{unsrtnat}
\bibliography{references}

\end{document}